\def\year{2021}\relax
\documentclass[letterpaper]{article} 
\usepackage{aaai21}  
\usepackage{times}  
\usepackage{helvet} 
\usepackage{courier}  
\usepackage[hyphens]{url}  
\usepackage{graphicx} 
\usepackage{graphicx}  
\usepackage{natbib}  
\usepackage{caption} 
\usepackage{amsmath}
\usepackage{bm}
\usepackage{multirow}
\usepackage[switch]{lineno}  %

\title{Zero-Shot Sketch-Based Image Retrieval with Structure-aware Asymmetric Disentanglement}
\author {
        Jiangtong Li,
        Zhixin Ling,
        Li Niu,
        Liqing Zhang \\
}
\affiliations {
    Shanghai Jiao Tong University \\
    {\tt\small jiangtongli1997@gmail.com}, \quad {\tt\small 1069066484@sjtu.edu.cn}, \\{\tt\small ustcnewly@gmail.com},\quad {\tt\small zhang-lq@cs.sjtu.edu.cn}
}
\begin{document}


\maketitle

\begin{abstract}

The goal of Sketch-Based Image Retrieval (SBIR) is using free-hand sketches to retrieve images of the same category from a natural image gallery. 
However, SBIR requires all test categories to be seen during training, which cannot be guaranteed in real-world applications. 
So we investigate more challenging Zero-Shot SBIR (ZS-SBIR), in which test categories do not appear in the training stage. 
After realizing that sketches mainly contain structure information while images contain additional appearance information, we attempt to achieve structure-aware retrieval via asymmetric disentanglement.
For this purpose, we propose our STRucture-aware Asymmetric Disentanglement (STRAD) method, in which image features are disentangled into structure features and appearance features while sketch features are only projected to structure space.  Through disentangling structure and appearance space, bi-directional domain translation is performed between the sketch domain and the image domain.
Extensive experiments demonstrate that our STRAD method remarkably outperforms state-of-the-art methods on three large-scale benchmark datasets.

\end{abstract}

\section{Introduction}

In recent years, with the rapid growth of multimedia data on the internet, image retrieval is playing a more and more important role in many fields.
Since sketches can be easily drawn and reveal some characteristics of the target images, sketch-based image retrieval (SBIR), which uses a sketch to retrieve the images of the same category, has become widely accepted among users. 
Therefore, SBIR has also attracted widespread attention in research community (\cite{saavedra2014sketch, saavedra2015sketch, li2016fine, sangkloy2016sketchy, lu2018learning}). 

\begin{figure}[ht]
\centering
\includegraphics[width=1\linewidth]{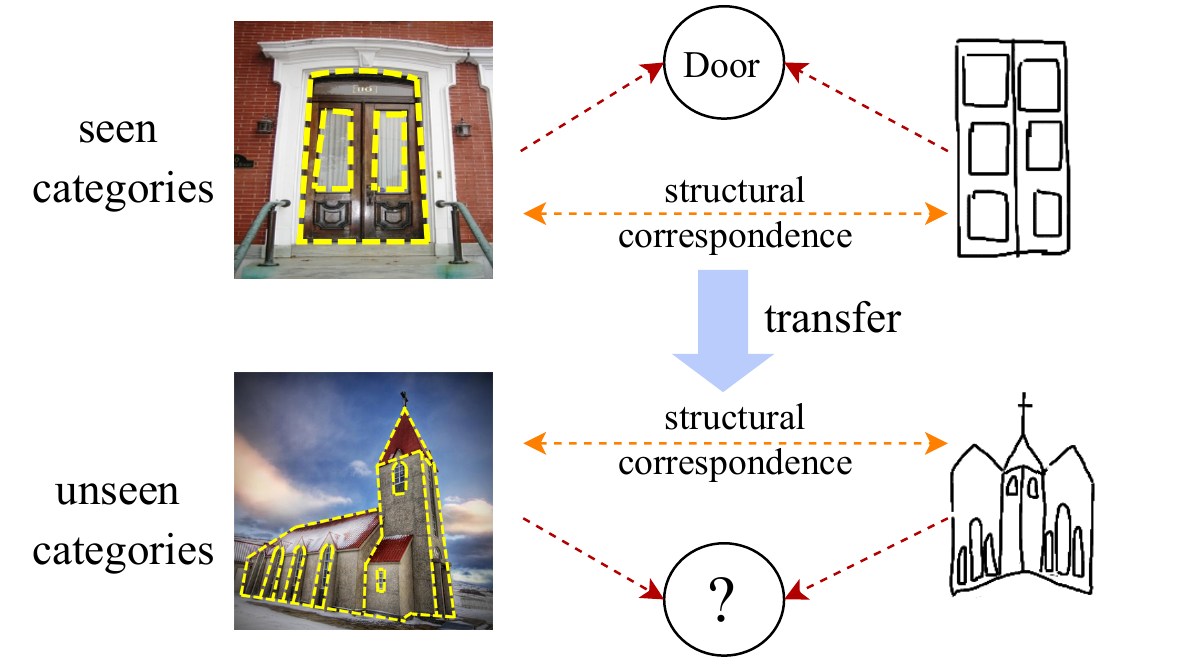}
   \caption{Traditional SBIR methods which correlate sketches/images with their category labels cannot generalize well to unseen categories. Our STRAD method learns structural correspondence between sketches and images from seen categories, which can be transferred to unseen ones.}
\label{fig:inshort}
\end{figure}

In the conventional SBIR setting, it assumes that the images and sketches in training and test sets share the same set of categories.
However, in real-world applications, the categories of test sketches/images may be out of the scope of training categories, leading to a more challenging task called zero-shot sketch-based image retrieval (ZS-SBIR) \cite{shen2018zero}, which assumes that test categories do not appear in the training stage. In the remainder of this paper, we refer to training  (\emph{resp.}, test) categories as seen (\emph{resp.}, unseen) categories \cite{dupont2018learning}.
Traditional SBIR methods~\cite{yu2017sketch} suffer from sharp performance drop in ZS-SBIR setting, because traditional SBIR methods may take a shortcut by correlating sketches/images with their category labels and retrieving the images from the same category as the query sketch~\cite{yelamarthi2018zero}. This shortcut is very effective when test data share the same categories as training data, but can hardly generalize to unseen categories as depicted in Figure~\ref{fig:inshort}.

To overcome the drawbacks of traditional SBIR methods in ZS-SBIR setting, several ZS-SBIR methods have been proposed to boost the performance on unseen categories, which can be categorized into the following groups:
1) \citet{yelamarthi2018zero} used a generative model based on aligned sketch-image pairs (a sketch is drawn based on a given image and thus has roughly the same outline as this image) to learn better correlation between sketches and images. However, the aligned sketch-image pairs are either unavailable or very expensive;
2) Some works \cite{shen2018zero, dutta2019semantically, verma2019generative, dey2019doodle, zhang2020zero} employed category-level semantic information to reduce the gap between seen categories and unseen categories. Whereas, category-level semantic information like word vectors \cite{mikolov2013distributed} are sometimes inaccessible or ambiguous. Besides, they only use semantic information of seen categories during training, which may limit the generalization ability to unseen categories;
3) \citet{liu2019semantic} located the catastrophic forgetting phenomenon and preserved the knowledge of model pretrained on ImageNet \cite{deng2009imagenet}. Despite its competitive performance, it relies on auxiliary WordNet \cite{miller1998wordnet} knowledge and its performance gain is mainly from the pre-training strategy;
4) \citet{dutta2019style} proposed to disentangle the representations of two domains (\emph{i.e.}, sketch and image) into domain-independent and domain-specific representations. Nevertheless, this symmetric disentanglement approach is not well-tailored for SBIR task and ignores the asymmetric relation between sketch and image domain, that is, sketches only contain structure information (\emph{e.g.}, outline, shape) while images additionally contain appearance information (\emph{e.g.}, color, texture, and background). 

Being aware of the asymmetric relation between sketch domain and image domain, we conjecture that the key of sketch-based image retrieval might be successfully matching the structure information between sketches and images. 
In this paper, we attempt to learn the structural correspondence between sketches and images on seen categories, which could generalize to unseen categories and facilitate ZS-SBIR task. 
For example, as shown in Figure~\ref{fig:inshort}, the structural correspondence between sketches and images within the same category can be learned based on seen categories, \emph{e.g.}, the structural similarity between ``door'' sketch and ``door'' image \emph{w.r.t.} the global/local contour and the layout of different components. In the test stage, given sketches and images from an unseen category ``church'', without knowing which category they belong to, we can still verify whether they belong to the same category on the premise of their structural similarity learnt from seen categories.

To obtain the structure features of sketches and images, we propose a STRucture-aware Asymmetric Disentanglement (STRAD) method to disentangle image feature into structure feature (\emph{e.g.}, outline, shape) and appearance feature (\emph{e.g.}, color, texture, and background) while projecting sketch feature into structure feature only. 
Our asymmetric disentanglement method is different from symmetric disentanglement in \citet{dutta2019style}, because our disentangled representations have explicit meanings (\emph{i.e.}, structure and appearance) and are specifically designed for SBIR task. 
Note that although several previous methods~\cite{yu2016sketch, liu2017deep} also intended to obtain the structure information of images by directly extracting edge maps from images, the low-level edge maps obtained in this way may not be very reliable due to possibly noisy and redundant information, which severely compromises the effectiveness of these methods in ZS-SBIR. 
In contrast, our method could extract high-level robust structure features from both sketches and images.

Our proposed STRAD method is illustrated in Figure \ref{fig:structure}. We first use a pre-trained model to extract features from sketches (\emph{resp.}, images) as sketch (\emph{resp.}, image) features. 
Then, the image features are disentangled into structure features and appearance features, while the sketch features are also projected to the structure space. 
Furthermore, bi-directional domain translation is performed through the structure features and appearance features.
Concretely, for image-to-sketch translation, we project the image features to structure features and then generate sketch features from the structure features. For sketch-to-image translation, we project the sketch features to structure features, which are combined with variational appearance features to compensate for the appearance uncertainty when generating image features from the sketch features.

Finally, we perform retrieval in all three spaces (\emph{i.e.}, structure space, image space, and sketch space), to combine the best of three worlds. 
Apparently, the retrieval in structure space and sketch space is structure-aware. Image feature is generated from structure feature and variational appearance feature. Since variational appearance feature is category-agnostic, the retrieval in image space is also structure-aware.
The effectiveness of our proposed STRAD method is verified by comprehensive experimental results on three large-scale benchmark datasets. Our main contributions are summarized as follows:
\begin{itemize}
    \item Based on the asymmetric relation between image domain and sketch domain, we propose to learn structural correspondence between these two domains via asymmetric disentanglement.
    \item We design a novel STRucture-aware Asymmetric Disentanglement (STRAD) method and a hybrid retrieval strategy in three spaces.
    \item Comprehensive results on three popular benchmark datasets show that our method significantly outperforms the state-of-the-art methods.
\end{itemize}

\section{Related Work}

\subsection{SBIR and ZS-SBIR}
The main goal of sketch-based image retrieval (SBIR) is to bridge the gap between image domain and sketch domain. 
Before deep learning was introduced to this task, hand-crafted based methods generally extracted the edge maps from natural images and then matched them with sketches using hand-craft feature \cite{eitz2010evaluation}. 
In recent years, deep learning based methods have become popular in this area. 
To reduce the gap between image domain and sketch domain, variants of siamese network \cite{qi2016sketch} and ranking loss \cite{sangkloy2016sketchy} were adopted for this task. Besides, semantic information and adversarial loss were also introduced to preserve the domain invariant information \cite{zhu2016learning,chen2018deep}.

Zero-shot sketch-based image retrieval (ZS-SBIR) was proposed by \citet{shen2018zero}. To reduce the intra-class variance in sketches and stabilize the training process, semantic information was leveraged in \cite{dutta2019semantically, verma2019generative, dey2019doodle, dutta2019style, zhang2020zero}. To reduce the gap between seen and unseen categories, a generative model along with aligned data pairs was proposed in \cite{yelamarthi2018zero}. To adapt the pre-trained model to ZS-SBIR without forgetting the knowledge from ImageNet, semantic-aware knowledge preservation was used in \citet{liu2019semantic}. However, all of the above methods did not consider the special relation between images and sketches, and treated them equally in their models.

\subsection{Disentangled Representation}
Disentangled representation learning aims to divide the latent representation into multiple units, with each unit corresponding to one latent factor (\emph{e.g.}, position, scale). 
Disentangled representations are more generalizable and semantically meaningful, and thus useful for a variety of tasks.

Disentangled representation learning methods can be categorized into unsupervised methods and supervised methods according to whether supervision for latent factors is available.
For unsupervised disentanglement, abundant methods \cite{chen2016infogan, higgins2017beta, chen2018isolating} have been developed. 
Most of them encouraged statistical independence across different dimensions of the latent representation while maintaining the mutual information between input data and latent representations.
For supervised disentanglement, \citet{kingma2014semi} used disentangled representation to enhance semi-supervised learning. \citet{zheng2019joint} proposed DG-Net to integrate discriminative and generative learning using disentangled representation. \citet{hadad2018two} proposed a two-step disentanglement method, which disentangles the label information from the original representation and enables feature reconstruction from decomposed features. Besides, supervised disentanglement has been applied to different tasks, like face recognition \cite{liu2018exploring}, image generation \cite{jha2018disentangling}, and style transfer \cite{yang2019tet}. 
Our work applies asymmetric disentangled representation learning to facilitate structure-aware retrieval.

\subsection{Domain Translation}
Many domain translation approaches, like CycleGAN \cite{zhu2017unpaired} and DiscoGAN \cite{kim2017learning}, have been proposed to translate figures between two domains (\emph{e.g.}, sketch and image domains).
In this subsection, we mainly discuss the domain translation methods based on disentangled representation. 
Overall speaking, they disentangle latent representation into domain-specific representation and domain-invariant representation. In our problem, structure (\emph{resp.}, appearance) features can be treated as domain-invariant (\emph{resp.}, specific) representation. The translation between two domains in previous works~\cite{lee2018diverse,huang2018multimodal} is generally symmetric. In contrast, the translation between sketch domain and image domain in our problem is asymmetric because image domain has additional domain-specific representation compared with sketch domain.

\begin{figure*}[ht]
\centering
\includegraphics[width=0.91\linewidth]{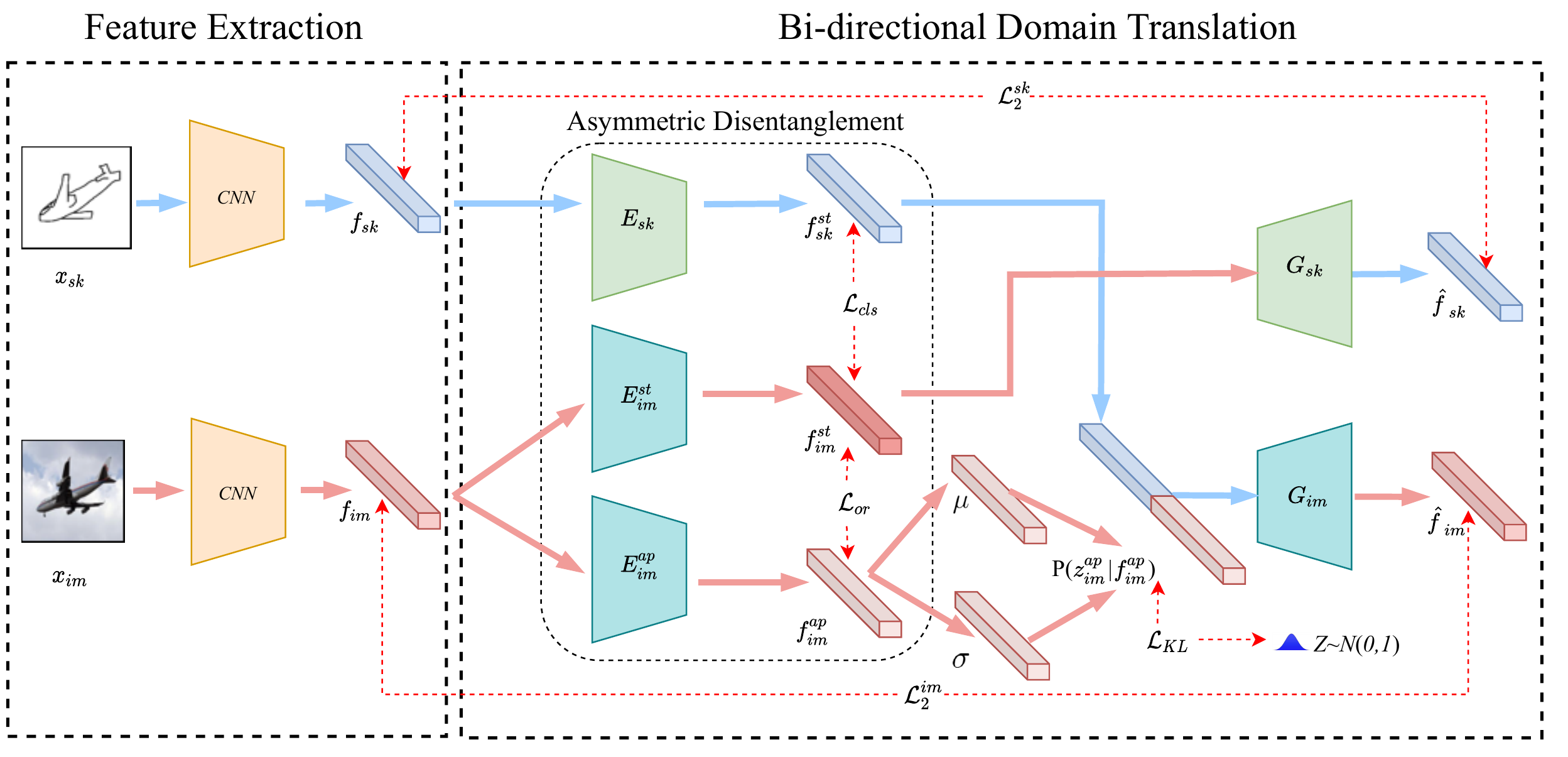}   
\caption{An overview of our STRAD method. We first adopt VGG-16 \cite{simonyan2014very} to extract features from images and sketch. Then we disentangle the image feature into appearance feature and structure feature, through which bi-directional domain translation is performed between the image domain and the sketch domain.}
\label{fig:structure}
\end{figure*}

\section{Methodology}

In this section, we introduce our proposed STRAD method for zero-shot sketch-based image retrieval. 
In Section~\ref{3.1}, we state the problem definition and notation. 
In Section~\ref{3.2}, we elaborate our STRAD method in detail. 
In Section~\ref{3.3}, we discuss our retrieval strategy.

\subsection{Problem Definition} \label{3.1}
In this paper, we focus on zero-shot sketch-based image retrieval, where only the sketches and images from seen categories are used for training. 
In the test stage, our proposed method is expected to use the sketches to retrieve the images, the categories of which are unseen during training.

Formally, given a sketch dataset $\mathcal{S}_{sk}$=$\{(\mathbf{x}_{sk,i}, y_i)|y_i \in \mathcal{Y}\}$ and an image dataset $\mathcal{S}_{im}$=$\{(\mathbf{x}_{im,j}, y_j) | y_j \in \mathcal{Y}\}$, where $\mathcal{Y}$ is category label set, and $(\mathbf{x}_{sk,i}, y_i)$ (\emph{resp.}, $(\mathbf{x}_{im,j}, y_j)$) represents a sketch (\emph{resp.}, image) with its corresponding category label, we follow the zero-shot setting in \citet{shen2018zero} to split all categories $\mathcal{Y}$ into $\mathcal{Y}^{tr}$ and $\mathcal{Y}^{te}$, in which no overlap exists between two label sets, \emph{i.e.}, $\mathcal{Y}^{tr} \cap \mathcal{Y}^{te} = \emptyset$. 
Based on the partition of label set $\mathcal{Y}$, we can split the sketch (\emph{resp.}, image) dataset into $\mathcal{S}^{tr}_{sk}$ and $\mathcal{S}^{te}_{sk}$ (\emph{resp.}, $\mathcal{S}^{tr}_{im}$ and $S^{te}_{im}$).
In the training stage, our model can only process the data in $\mathcal{S}^{tr}_{sk}$ and $\mathcal{S}_{im}^{tr}$. 
During testing, given a sketch $x_{sk}$ from $\mathcal{S}^{te}_{sk}$, our model needs to retrieve the images belonging to the same category from test image gallery $\mathcal{S}^{te}_{im}$.

\subsection{Our Framework} \label{3.2}

An overview of our method is illustrated in Figure \ref{fig:structure}. In this section, we will introduce our method from Asymmetric Disentanglement and Bi-direction Domain Translation.

\subsubsection{Asymmetric Disentanglement}
To achieve the goal of structure-aware retrieval, this module should decompose structure information and appearance information, and align images and sketches in the shared structure space. 

Therefore, we first disentangle the image features into appearance features and structure features.
Given an image $\mathbf{x}_{im}$ and a sketch $\mathbf{x}_{sk}$ from the same category $y$ , we first use fixed backbone model, \emph{i.e.}, VGG-16 \cite{simonyan2014very}, to produce their image feature $\mathbf{f}_{im}$ and sketch feature $\mathbf{f}_{sk}$.
For image feature $\mathbf{f}_{im}$, we adopt two image encoders $E_{im}^{st}$ and $E_{im}^{ap}$ to disentangle $\mathbf{f}_{im}$ into image structure feature $\mathbf{f}_{im}^{st}$ and image appearance feature $\mathbf{f}_{im}^{ap}$. Besides, to project sketch feature $\mathbf{f}_{sk}$ to the same structure space as $\mathbf{f}_{im}^{st}$, a sketch encoder $E_{sk}^{st}$ is adopted to obtain sketch structure feature $\mathbf{f}_{sk}^{st}$. The above process is formulated as follows,
\begin{equation}
    \mathbf{f}_{im}^{ap} \!=\! E_{im}^{ap}(\mathbf{f}_{im}); \quad
    \mathbf{f}_{im}^{st} \!=\! E_{im}^{st}(\mathbf{f}_{im}); \quad
    \mathbf{f}_{sk}^{st} \!=\! E_{sk}^{st}(\mathbf{f}_{sk}).
\end{equation}

To capture the structural correspondence between images and sketches, we expect the structure features from both images and sketches to be aligned in the same space. 
Moreover, in the structure space shared by sketch and image, we expect intra-class coherence and inter-class separability across different domains (\emph{i.e.}, sketch domain and image domain). 
Specifically, we expect to pull close the image/sketch structure features of the same category and push apart the image/sketch structure features from different categories. 
It has been proved in \citet{liu2019semantic} that a simple classification loss can accomplish the above task well. Therefore, we employ a structure classifier on both image structure features and sketch structure features to distinguish their category labels, by using the cross-entropy classification loss:
\begin{equation} \label{eqn:cls_loss}
\begin{aligned}
    \mathcal{L}_{cls} = &-\log \frac{\exp({\bm{w}_{y}^{\text{T}}}\mathbf{f}_{im}^{st}+b_{y})}{\sum_{k\in \mathcal{Y}^{tr}}\exp({\bm{w}_{k}^{\text{T}}\mathbf{f}_{im}^{st}+b_{k}})} \\
    &-\log \frac{\exp({\bm{w}_{y}^{\text{T}}}\mathbf{f}_{sk}^{st}+b_{y})}{\sum_{k\in \mathcal{Y}^{tr}}\exp({\bm{w}_{k}^{\text{T}}\mathbf{f}_{sk}^{st}+b_{k}})},
\end{aligned}
\end{equation}
where $\bm{w}_k$ and $b_k$ are learnable parameters in the structure classifier corresponding to category $k$. Recall that $y$ is the category label of $\mathbf{x}_{im}$ and $\mathbf{x}_{sk}$.
Although the structure space regulated by (\ref{eqn:cls_loss}) seems sufficient to perform sketch-based image retrieval, the following modules to be introduced can further help capture the structural correspondence and improve the generalization ability (see experiments in Section\ref{4.3} and \ref{4.5}).

After enforcing the structure features of sketches and images to the same structure space, we further expect that the appearance features of images only contain complementary information (\emph{e.g.}, color, texture, and background) to the structure features.
To reinforce the disentanglement of image features, we impose an orthogonal constraint between structure and appearance features of images based on cosine similarity \cite{shukla2019product}:
\begin{equation} \label{eqn:orth_loss}
    \mathcal{L}_{or} = \cos(\mathbf{f}_{im}^{ap}, \mathbf{f}_{im}^{st}) = \frac{\mathbf{f}_{im}^{ap} \cdot \mathbf{f}_{im}^{st}}{||\mathbf{f}_{im}^{ap}||_2 ||\mathbf{f}_{im}^{st}||_2},
\end{equation}
where $\cdot$ means the the dot product between two vectors.
Note that $\mathbf{f}_{im}^{ap}$ and $\mathbf{f}_{im}^{st}$ are the output of ReLU activation, so $\cos(\mathbf{f}_{im}^{ap}, \mathbf{f}_{im}^{st})$ is always non-negative and minimizing Equ. (\ref{eqn:orth_loss}) will push $\cos(\mathbf{f}_{im}^{ap}, \mathbf{f}_{im}^{st})$ towards zero.

\subsubsection{Bi-directional Domain Translation}

To learn better disentangled representations and fully utilize the disentangled image features, we perform bi-directional domain translation between the sketch domain and the image domain.

For image-to-sketch translation, we aim to translate image feature $\mathbf{f}_{im}$ to sketch feature through its image structure feature $\mathbf{f}_{im}^{st}$. We employ a decoder $G_{sk}$ to generate sketch feature  $\hat{\mathbf{f}}_{sk}$ from $\mathbf{f}_{im}^{st}$ by $\hat{\mathbf{f}}_{sk}= G_{sk}(\mathbf{f}_{im}^{st})$. Considering that $\hat{\mathbf{f}}_{sk}$ and $\mathbf{f}_{sk}$ belong to the same category, we enforce the generated sketch features to be close to the real sketch features from the same category by using $L_2$ loss:
\begin{equation} \label{eqn:resk}
    \mathcal{L}_{2}^{sk} = ||\mathbf{f}_{sk}-\hat{\mathbf{f}}_{sk}||_2.
\end{equation}

For sketch-to-image translation, we aim to translate sketch feature $\mathbf{f}_{st}$ to image feature through its sketch structure feature $\mathbf{f}_{sk}^{st}$.
However, images contain extra appearance information (\emph{e.g.}, color, texture, and background) compared with sketches, so it is necessary to compensate for the appearance uncertainty when translating from structure features to image features. 
Therefore, in the training stage, appearance feature $\mathbf{f}_{im}^{ap}$ could be integrated with sketch structure feature $\mathbf{f}_{sk}^{st}$ to generate the image feature. 

During testing, given a sketch, we also hope to generate its image feature to enable retrieval in the image space. 
Nevertheless, we do not have the corresponding appearance feature. 
A common solution is stochastic sampling. 
We introduce a variational estimator $V_{im}^{ap}$ to approximate the variational Gaussian distribution $P(\mathbf{z}_{im}^{ap}|\mathbf{f}_{im}^{ap})$ based on $\mathbf{f}_{im}^{ap}$, that is, $(\bm{\mu}_{im}^{ap}, \bm{\sigma}_{im}^{ap})= V_{im}^{ap}(\mathbf{f}_{im}^{ap})$. 
Then, we use Kullback-Leibler divergence to enforce  $P(\mathbf{z}_{im}^{ap}|\mathbf{f}_{im}^{ap})$ to be close to prior distribution $\mathcal{N}(\mathbf{0}, \mathbf{1})$ to support stochastic sampling:
\begin{equation} \label{eqn:kl}
    \mathcal{L}_{KL} = KL(\mathcal{N}(\bm{\mu}_{im}^{ap}, \bm{\sigma}_{im}^{ap}) || \mathcal{N}(\mathbf{0}, \mathbf{1})).
\end{equation}

After using reparameterization trick \cite{kingma2013auto} to sample variational appearance feature $\mathbf{z}_{im}^{ap}$, \emph{i.e.}, $\mathbf{z}_{im}^{ap}=\bm{\mu}_{im}^{ap}+\epsilon \bm{\sigma}_{im}^{ap}$, where $\epsilon$ is sampled from $\mathcal{N}(0,1)$, we employ a decoder $G_{im}$ to generate $\hat{\mathbf{f}}_{im} = G_{im}([\mathbf{z}_{im}^{ap}, \mathbf{f}_{sk}^{st}])$ based on $\mathbf{z}_{im}^{ap}$ and $\mathbf{f}_{sk}^{st}$, where $[\cdot,\cdot]$ means concatenating two vectors. Considering that $\hat{\mathbf{f}}_{im}$ has the same category label as $\mathbf{f}_{im}$ and its appearance uncertainty comes from $\mathbf{f}_{im}$, we enforce $\hat{\mathbf{f}}_{im}$ to be close to $\mathbf{f}_{im}$ with  $L_2$ loss:
\begin{equation} \label{eqn:reim}
    \mathcal{L}_{2}^{im} = ||\mathbf{f}_{im}-\hat{\mathbf{f}}_{im}||_2.
\end{equation}

By performing image-to-sketch translation, we expect that the image structure features contain the necessary structure information to generate sketch features. By performing sketch-to-image translation, we expect that the appearance features contain the necessary information of appearance uncertainty to compliment the sketch structure features when generating image features. Therefore, bi-directional domain translation could cooperate with classification loss and orthogonal loss to assist feature disentanglement.

Finally, the full objective function can be expressed as
\begin{equation} \label{eqn:total_loss}
    \mathcal{L} \!=\! \mathcal{L}_{or} \!+\! \mathcal{L}_{cls} \!+\! \mathcal{L}_{KL} \!+\!  \mathcal{L}_{2}^{im} \!+\! \mathcal{L}_{2}^{sk}.
\end{equation}

\subsection{Retrieval Strategy} \label{3.3}

In the test stage, we perform retrieval in all three spaces: structure, sketch, and image spaces. Specifically, given a sketch $\mathbf{x}_{sk}$ with sketch feature $\mathbf{f}_{sk}$  and an image $\mathbf{x}_{im}$ with image feature $\mathbf{f}_{im}$ , we compare them in three spaces to combine the best of all worlds.

\noindent (1) \textbf{Structure space}: We project $\mathbf{f}_{im}$ and $\mathbf{f}_{sk}$ into the common structure space by $\mathbf{f}_{im}^{st}=E_{im}^{st}(\mathbf{f}_{im})$ and $\mathbf{f}_{sk}^{st}=E_{sk}^{st}(\mathbf{f}_{sk})$ respectively, so we can calculate the distance in structure space $\mathcal{D}_{st} =1-cos(\mathbf{f}_{im}^{st}, \mathbf{f}_{sk}^{st})$.

\noindent (2) \textbf{Sketch space}: For image $\mathbf{x}_{im}$, based on its image structure feature $\mathbf{f}_{im}^{st}$, we employ the sketch decoder $G_{sk}$ to generate its sketch feature $\hat{\mathbf{f}}_{sk}=G_{sk}(\mathbf{f}_{im}^{st})$, so we can calculate the distance in sketch space $\mathcal{D}_{sk} =1-cos(\hat{\mathbf{f}}_{sk}, \mathbf{f}_{sk})$.

\noindent (3) \textbf{Image space}: For sketch $\mathbf{x}_{sk}$, based on its sketch structure feature $\mathbf{f}_{sk}^{st}$ and a variational appearance feature $\mathbf{z}_i$ sampled from $\mathcal{N}(\mathbf{0},\mathbf{1})$, we employ the image decoder $G_{im}$ to generate its image feature. We can generate $N$ image features vectors by sampling $N$ times and use the average to represent the final image feature $\hat{\mathbf{f}}_{im}$:
    \begin{equation}
        \hat{\mathbf{f}}_{im} = \frac{1}{N}\sum_{i=1}^{N}G_{im}([\mathbf{z}_i,\mathbf{f}_{sk}^{st}]).
    \end{equation}
    
So we can calculate the distance in image space $\mathcal{D}_{im} =1-cos(\mathbf{f}_{im}, \hat{\mathbf{f}}_{im})$. Finally, we calculate the weighted average of three distances for the best retrieval:
\begin{equation}
\begin{aligned}
    \mathcal{D}_{fusion} = \lambda_1(\mathcal{D}_{im} +  \mathcal{D}_{sk})
     +\lambda_2 \mathcal{D}_{st}  ,
\end{aligned}
\end{equation}
where  $\lambda_1$ and $\lambda_2$ are hyper-parameters to balance different spaces. Since sketch feature is generated solely based on structure feature, retrieval in sketch space is obviously structure-aware. Image feature is generated based on structure feature and variational appearance feature $\mathbf{z}_i$, but $\mathbf{z}_i$ is sampled from $\mathcal{N}(\mathbf{0},\mathbf{1})$ for any category and thus category-agnostic. Therefore, retrieval in image space mainly depends on the structure information and is also structure-aware.

\begin{table*}[ht]
\centering
\resizebox{0.95\textwidth}{!}{%
\begin{tabular}{cccccccc}
\hline\hline
\multirow{2}{*}{} & \multirow{2}{*}{Method} & \multicolumn{2}{c}{Sketchy Ext.} & \multicolumn{2}{c}{TU-Berlin Ext.} & \multicolumn{2}{c}{QuickDraw Ext.} \\ \cline{3-8} 
 &  & P@200(\%) & mAP@200(\%) & P@200(\%) & mAP@200(\%) & P@200(\%) & mAP@200(\%) \\ \hline
\multirow{2}{*}{SBIR} & SaN \cite{yu2017sketch} & 15.3$\pm$0.6 & 5.8$\pm$0.4 & 10.1$\pm$0.6 & 4.2$\pm$0.4 & 4.2$\pm$0.5 & 0.9$\pm$0.2 \\
 & Siamese \cite{qi2016sketch} & 25.6$\pm$0.6 & 15.3$\pm$0.5 & 8.3$\pm$0.4 & 3.7$\pm$0.3 & 4.0$\pm$0.3 & 0.7$\pm$0.2 \\ \hline
\multirow{4}{*}{ZSL} & ESZSL \cite{romera2015embarrassingly} & 20.9 & 11.8 & 8.5 & 3.4 & 6.3 & 1.8 \\
 & SAE \cite{kodirov2017semantic} & 26.1 & 14.5 & 10.4 & 4.6 & 6.8 & 1.9 \\
 & CMT \cite{socher2013zero} & 27.3$\pm$0.7 & 15.8$\pm$0.5 & 10.8$\pm$0.5 & 4.9$\pm$0.4 & 6.3$\pm$0.4 & 1.7$\pm$0.3 \\
 & DeViSE \cite{frome2013devise} & 18.2$\pm$0.5 & 7.2$\pm$0.4 & 7.3$\pm$0.2 & 2.2$\pm$0.2 & 6.7$\pm$0.2 & 2.6$\pm$0.1 \\ \hline
\multirow{8}{*}{ZS-SBIR (D)} & Simple DB & 32.1$\pm$1.3 & 19.0$\pm$0.9 & 14.6$\pm$1.2 & 6.8$\pm$0.8 & 7.1$\pm$0.6 & 3.1$\pm$0.5 \\
 & CVAE \cite{yelamarthi2018zero} & 39.3$\pm$0.9 & 26.3$\pm$0.8 & 15.2$\pm$0.7 & 7.5$\pm$0.5 & 6.4$\pm$0.4 & 1.8$\pm$0.3 \\
 & \citet{xu2019semantic} & 42.8$\pm$1.1 & 31.1$\pm$0.8 & 17.1$\pm$0.7 & 9.5$\pm$0.5 & 7.2$\pm$0.6 & 3.0$\pm$0.3 \\
 & SEM-PCYC \cite{dutta2019semantically} & 43.8$\pm$0.7 & 31.6$\pm$0.5 & 19.5$\pm$0.8 & 10.7$\pm$0.4& 9.4$\pm$0.4 & 3.7$\pm$0.2 \\
 & Doodle \cite{dey2019doodle} & 43.2$\pm$0.8 & 30.1$\pm$0.5 & 19.3$\pm$1.1 & 10.2$\pm$0.7 & 9.8$\pm$0.6 & 3.7$\pm$0.4 \\
 & \citet{dutta2019style} & 43.0$\pm$0.7 & 28.5$\pm$0.4 & 17.8$\pm$0.5 & 10.8$\pm$0.3 & 8.3$\pm$0.3 & 3.5$\pm$0.1 \\
 & SketchGCN \cite{zhang2020zero} & 45.1$\pm$1.3 & 34.8$\pm$0.9 & 21.3$\pm$0.7 & 12.8$\pm$0.5 & 9.7$\pm$0.6 & 4.0$\pm$0.4 \\
 & \textbf{STRAD} & \textbf{50.2}$\pm$0.7 & \textbf{37.9}$\pm$0.4 & \textbf{24.5}$\pm$0.9 & \textbf{15.4}$\pm$0.6 & \textbf{12.6}$\pm$0.5 & \textbf{5.4}$\pm$0.4 \\ \hline
\multirow{3}{*}{ZS-SBIR (S)} & Simple SB & 49.7$\pm$1.9 & 36.7$\pm$1.5 & 26.2$\pm$1.6 & 16.2$\pm$1.1 & 16.6$\pm$1.3 & 7.1$\pm$0.8 \\
 & SAKE \cite{liu2019semantic} & 51.9$\pm$0.7 & 39.4$\pm$0.5 & 28.7$\pm$0.6 & 18.1$\pm$0.4 & 18.4$\pm$0.5 & 8.4$\pm$0.3 \\
 & \textbf{STRAD} & \textbf{53.7}$\pm$0.8 & \textbf{41.3}$\pm$0.6 & \textbf{30.1}$\pm$0.7 & \textbf{19.9}$\pm$0.4 & \textbf{19.1}$\pm$0.4 & \textbf{8.6}$\pm$0.2 \\ \hline\hline
\end{tabular}%
}
\caption{Comparison of our STRAD method and baselines on Sketchy, TU-Berlin, and QuickDraw datasets. (D) is short for ``double backbone'' setting and (S) is short for ``single backbone'' setting. Best results are denoted in boldface in both settings.}
\label{tab:1}
\end{table*}

\section{Experiment}


\subsection{Dataset}
We evaluate our STRAD method and all the other baselines on three large-scale benchmark datasets: Sketchy \cite{sangkloy2016sketchy}, TU-Berlin \cite{eitz2012hdhso}, and QuickDraw \cite{dey2019doodle}. Following the same setting in \citet{shen2018zero}, Sketchy and TU-Berlin are extended with images obtained from \citet{liu2017deep}. Due to the limitation of space, the details of datasets, seen/unseen category split, and implementation are left to Supplementary Material.

\subsection{Comparison with Existing Methods}

We compare our model with 13 prior methods, which can be categorized into three categories: sketch-based image retrieval (SBIR), zero-shot learning (ZSL), and zero-shot SBIR (ZS-SBIR). 
The SBIR baselines include Siamese \cite{qi2016sketch}, and SaN \cite{yu2017sketch}.
The ZSL baselines include ESZSL \cite{romera2015embarrassingly}, SAE \cite{kodirov2017semantic}, CMT \cite{socher2013zero}, and DeViSE \cite{frome2013devise}.
For a fair comparison, we use fine-tuned VGG-16 as backbone for all methods except SaN, which specifically designs a  backbone for SBIR.
Among all the previous works on ZS-SBIR, either two different backbones or a single backbone can be used to extract image and sketch features. In this paper, we conduct experiments in both settings.
For the ``double backbone'' setting, we compare with CVAE \cite{yelamarthi2018zero}, \citet{xu2019semantic}, SEM-PCYC \cite{dutta2019semantically}, Doodle \cite{dey2019doodle}, \citet{dutta2019style}, and SketchGCN \cite{zhang2020zero}. 
For the ``single backbone'' setting, we compare with SAKE \cite{liu2019semantic} because they use single backbone in their paper.
During training, we fix the backbone for all methods except \citet{dey2019doodle} and SAKE~\cite{liu2019semantic}.
For \citet{dey2019doodle}, we strictly follow their paper to train the backbone during the training.
For SAKE \cite{liu2019semantic}, we strictly follow the training strategy introduced in their paper \footnote{Since SAKE~\cite{liu2019semantic} starts from backbone model pre-trained on ImageNet-1k to prevent knowledge forgetting, we do not change their setting.}.
Following \citet{yelamarthi2018zero}, we use mean average precision and precision of top 200 retrievals (mAP@200 and P@200) as the evaluation metrics. Due to the calculation of mAP@200 in \cite{dey2019doodle, zhang2020zero} is slightly different from ours, their reported mAP@200 are different from our reproduced results. 
We also report the performance of fine-tuned single backbone (\emph{resp.}, double backbones) as ``Simple SB'' (\emph{resp.}, ``Double DB'') in Table~\ref{tab:1}, where cosine distance between the image and sketch features are used for retrieval.
We run each model for 10 times with random seeds 1 - 10 and report the ``Mean $\pm$ Std'' results in Table~\ref{tab:1}, except for ESZSL and SAE, which do not introduce randomness

Based on Table~\ref{tab:1}, we find all SBIR and ZSL baselines underperform the ZS-SBIR baselines due to their poor generalization ability from seen categories to unseen ones.
On the TU-Berlin dataset, the results of several methods~\cite{liu2019semantic, dutta2019semantically,dutta2019style,zhang2020zero} are worse than those reported in their papers due to different seen/unseen category splits. 
In particular, the number of unseen categories under our split is twice larger than that in  \cite{liu2019semantic,zhang2020zero}. Our split criterion also prevents information leakage from ImageNet-1k to unseen categories.
The overall results on TU-Berlin are lower than those on Sketchy due to larger number of unseen categories (64 \emph{v.s.} 21). Furthermore, the overall results on TU-Berlin are lower than those on QuickDraw since sketches of QuickDraw were drawn by amateurs.

In ``double backbone'' setting, our proposed STRAD excels the state-of-the-art methods by 5.1\% on Sketchy, 3.3\% on TU-Berlin, and 2.8\% on QuickDraw in terms of P@200. In ``single backbone'' setting, we find that the ``Simple SB''  outperforms most methods in ``double backbone'' setting, which reveals that it might be the best solution for SBIR task to use a single model to pull close the image and sketch space. Starting from ``Simple SB'', the performance gain of our STRAD is smaller than that in ``double backbone'' setting, because a single backbone has already filtered out most differences between features of images and sketches. However, there still remains extra appearance information in image features, so our STRAD also outperforms SAKE \cite{liu2019semantic} and achieves the best results on all datasets.

To further demonstrate the effectiveness of our method, we also report the results on TU-Berlin following the splits in \citet{liu2019semantic} and the results in generalized ZS-SBIR setting \cite{dutta2019semantically} in Supplementary Material.

\begin{figure*}[t]
\begin{center}
\includegraphics[width=0.87\linewidth]{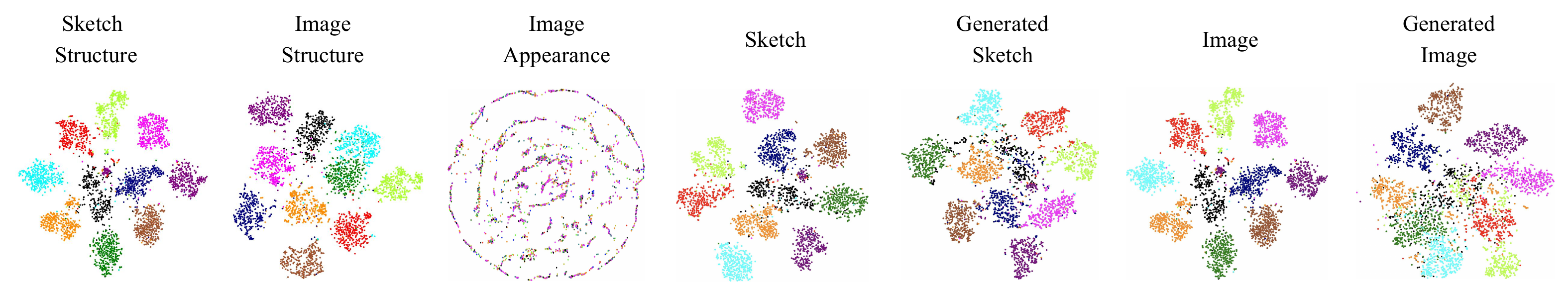}
\end{center}
   \caption{The t-SNE visualization of seven types of features on Sketchy test set. Best viewed in color.}
\label{fig:tsne}
\end{figure*}

\begin{figure}[ht]
\begin{center}
\includegraphics[width=0.87\linewidth]{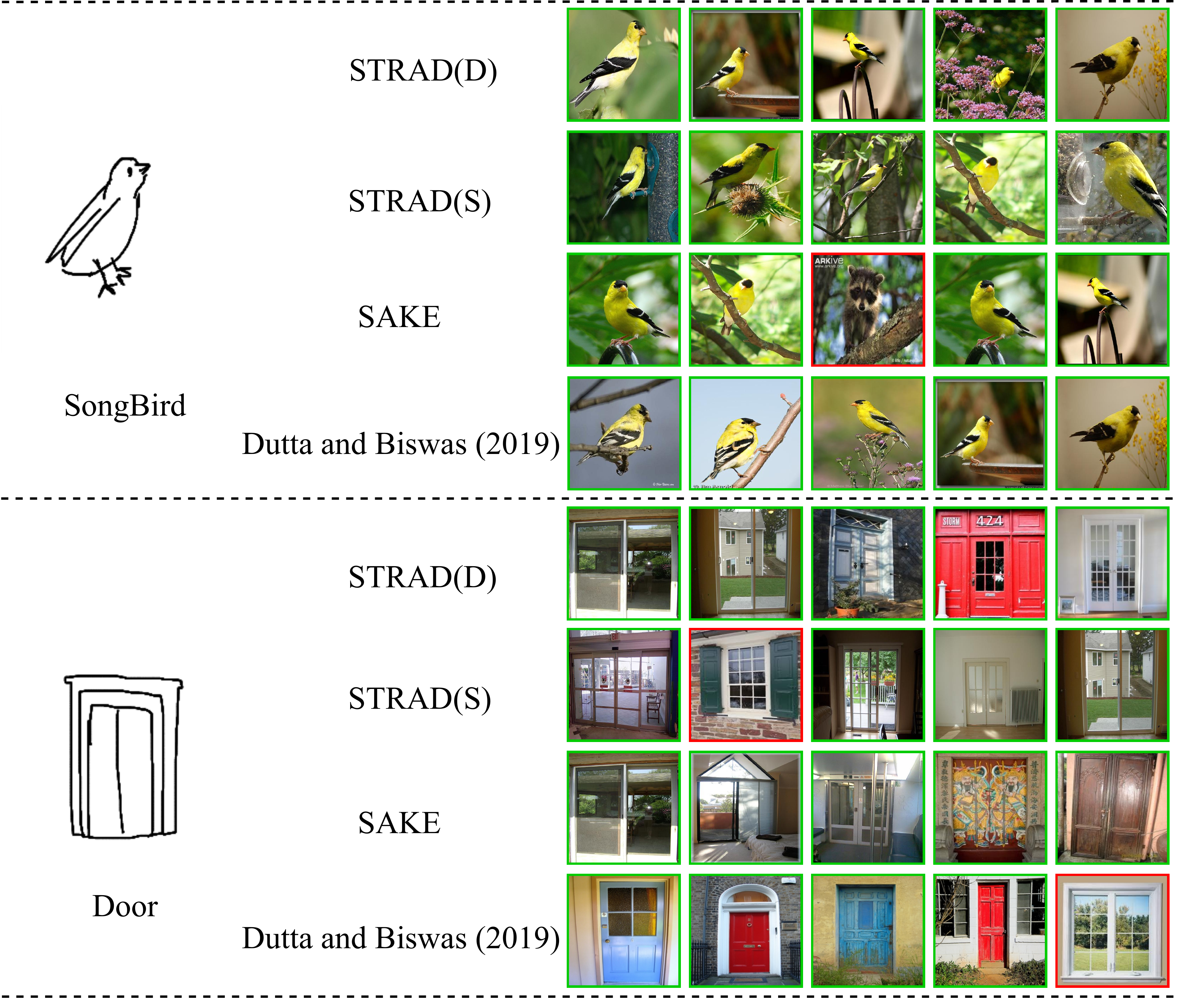}
\end{center}
   \caption{Top-5 images retrieved by STRAD (double backbone), STRAD (single backbone), SAKE\cite{liu2019semantic}, and \citet{dutta2019style} methods on Sketchy. The green (\emph{resp.}, red) border indicates the correct (\emph{resp.}, incorrect) retrieval results.}
\label{fig:case}
\end{figure}


\subsection{Ablation Study}\label{4.3}
By taking the Sketchy as an example, we analyze the effectiveness of different loss terms in ``double backbone'' setting. 
We ablate each loss term in Equ. (\ref{eqn:cls_loss}), (\ref{eqn:orth_loss}), (\ref{eqn:resk}), (\ref{eqn:kl}), and (\ref{eqn:reim}), and report the results in Table \ref{tab:3}. The random seed in this subsection is set as 1.

\begin{table}[ht]
\resizebox{0.45\textwidth}{!}{%
\begin{tabular}{cccccccccc}
\hline
& $\mathcal{L}_{KL}$ & $\mathcal{L}_{2}^{sk}$ & $\mathcal{L}_{2}^{im}$ & $\mathcal{L}_{cls}$ & $\mathcal{L}_{or}$ & Comb & Sk & Im & St \\ \hline
1 & $\surd$ & $\surd$ & $\surd$ & $\surd$ & $\surd$ & 50.2 & 46.9 & 43.8 & 47.1 \\
2 & $\times$ & $\surd$ & $\surd$ & $\surd$ & $\surd$ & 48.7 & 45.3 & 36.8 & 47.0 \\
3 & $\surd$ & $\times$ & $\surd$ & $\surd$ & $\surd$ & 47.2 & - & 42.7 & 46.1 \\
4 & $\times$ & $\surd$ & $\times$ & $\surd$ & $\surd$ & 48.4 & 46.5 & - & 45.9 \\
5 & $\surd$ & $\surd$ & $\surd$ & $\times$ & $\surd$ & 44.2 & 42.8 & 39.5 & - \\
6 & $\surd$ & $\surd$ & $\surd$ & $\surd$ & $\times$ & 49.4 & 46.8 & 43.2 & 46.9 \\
7 & $\times$ & $\times$ & $\times$ & $\surd$ & $\times$ & 41.9 & - & - & 41.9 \\ \hline
\end{tabular}%
}
\caption{The ablation studies results (P@200(\%)) of our STRAD method on Sketchy. Comb denotes the retrieval in the combination of all spaces (see Sec.~\ref{3.3}); Sk, Im, and St denotes the retrieval in sketch, image, and structure space individually.  $\surd$ (\emph{resp.}, $\times$) means adding (\emph{resp.}, removing) this loss during training, and $-$ means unavailable entries.}

\label{tab:3}
\end{table}

We observe that the performance in both image and sketch spaces drops significantly without the classification loss (``Im'' in row 1 \emph{v.s.} ``Im'' in row 5, and ``Sk'' in row 1 \emph{v.s.} ``Sk'' in row 5), which shows the significant contribution of classification loss to  feature disentanglement and  alignment in all three spaces.
In the meanwhile, only using classification loss leads to a performance drop in structure space (``St'' in row 1 \emph{v.s.} ``St'' in row 7).
Although the classification loss can pull close the sketch and image spaces, there is no guarantee that the structural correspondence between sketches and images is well captured. 
In contrast, asymmetric disentanglement and bi-directional domain translation in our full-fledged method can help capture the structural correspondence and generalize to unseen categories.
By comparing row 1 and row 3, we find that $\mathcal{L}_2^{sk}$ is also very important, because sketch features without involving appearance uncertainty are easier to be regularized than image features, which makes sketch space more crucial for the combination result.
Another observation is that the orthogonal loss which reinforces disentanglement does benefit the retrieval task (row 1 \emph{v.s.}  row 6). 


\subsection{Feature Space Analysis}

To demonstrate the ability of our model to disentangle the image features, 
in Figure \ref{fig:tsne}, we visualize 7 types of features from 10 unseen categories on Sketchy using t-SNE \cite{maaten2008visualizing}, including image appearance feature, image structure feature, sketch structure feature, image feature, generated image feature, sketch feature, and generated sketch feature. According to Figure \ref{fig:tsne}, we have the following observations: 1) Different categories can be separated well in ``image structure'' and ``sketch structure'', which significantly facilitates the retrieval in structure space; 2) The results in ``image structure'' and ``image appearance'' are complementary, in accordance with the disentanglement between structure and appearance features; 3) The results in ``sketch features'' and ``generated sketch features'' are similar, compared with those in ``image features'' and ``generated image features'', which implies that the generated sketch features could be closer to the real sketch features from the same category; 4) Compared results in ``generated sketch feature'', ``sketch structure'' and ``image structure'', the results in ``generated image feature'' show relatively poor separability, which limits retrieval performance in image space. 

\begin{figure}[ht]
\begin{center}
\includegraphics[width=0.87\linewidth]{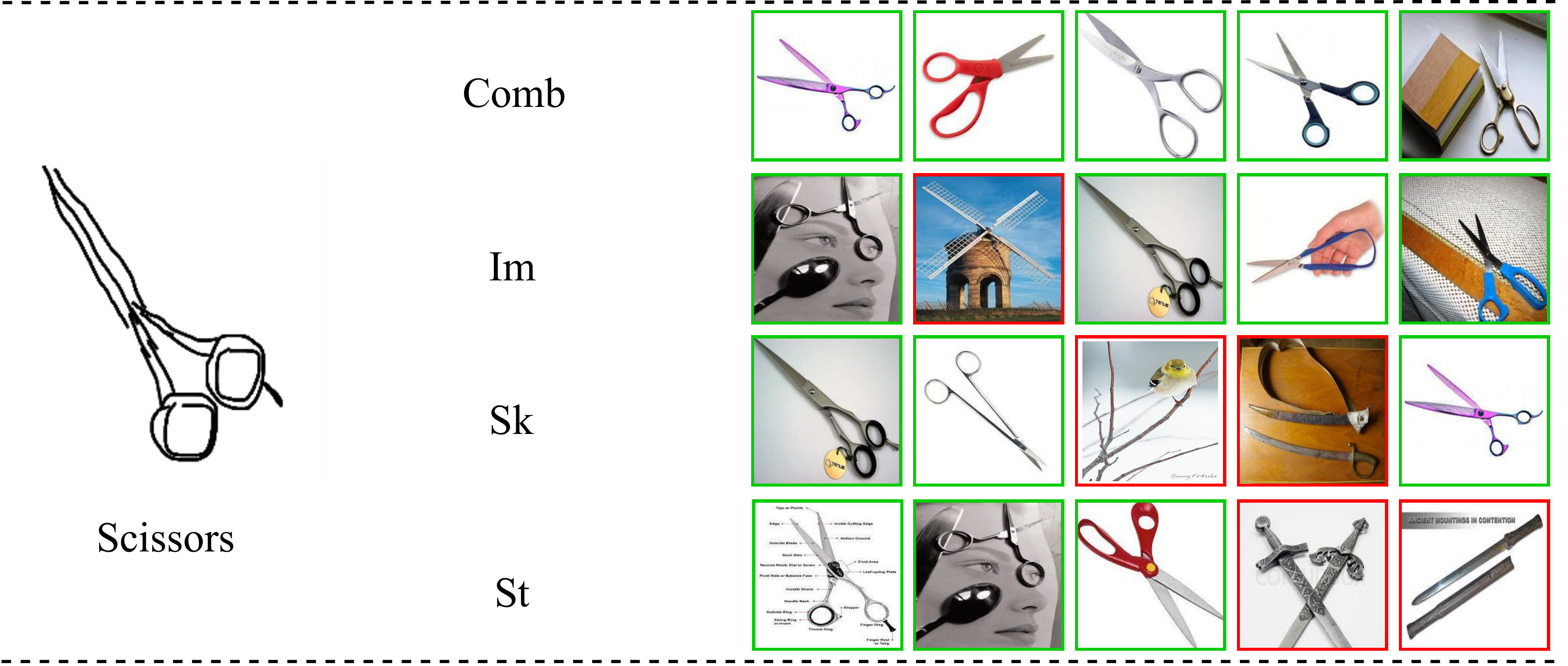}
\end{center}
   \caption{Top-5 images retrieved in combined, image, sketch, and structure space under double backbone setting. The green (\emph{resp.}, red) border indicates the correct (\emph{resp.}, incorrect) retrieval results.}
\label{fig:case2}
\end{figure}

\subsection{Case Study} \label{4.5}
In Figure \ref{fig:case}, we show top-5 retrieval results of STRAD (double backbone), STRAD (single backbone), SAKE~\cite{liu2019semantic}, and \citet{dutta2019style}, based on which we have the following observations. 
1)  Our STRAD is adept at capturing the correspondence between the retrieved images and the given sketch \emph{w.r.t.} both local structure information (\emph{e.g.}, door-case) and global structure information (\emph{e.g.}, global grid structure of door). Besides, in both ``single backbone'' and ``double backbone'' settings, our STRAD is able to retrieve the images with cluttered background (\emph{e.g.}, a bird with intricate leaves/flowers behind), which benefits from our structure-aware retrieval in three spaces. 
The above advantages come from the combination of three retrieval spaces. Detailed analyses and examples of complementarity of three retrieval spaces are provided in Supplementary.
2) For baselines,  SAKE \cite{liu2019semantic} with single backbone can retrieve images with cluttered background while \citet{dutta2019style} with double backbones can barely tolerate the complex backgrounds.
One possible reason is that using the same backbone for the sketch domain and image domain is to share a large amount of parameters for feature extraction, which would eliminate the background differences between these two domains.

In Figure \ref{fig:case2}, we show top-5 images retrieved in combined, image, sketch, and structure space by our STRAD (double backbone). It can be seen that image space tends to retrieve images with cluttered background while sketch and structure spaces are prone to retrieve images with clean background, which proves that image space contains extra appearance information (\emph{e.g.}, background) complemented by variational appearance feature. 
More examples and analyses of three retrieval spaces can be found in Supplementary Materials. 

\section{Conclusion}
In this work, we have studied the problem zero-shot sketch-based image retrieval (ZS-SBIR) from a new viewpoint, \emph{i.e.}, using asymmetric disentangled representation to facilitate structure-aware retrieval. We have proposed our STRucture-aware Asymmetric Disentanglement (STRAD) model, with retrieval performed in combination of three complementary spaces. Comprehensive experiments on three large-scale benchmark datasets have demonstrated the generalization ability of our model from seen categories to unseen ones.

\bibliography{STRAD}

\end{document}